# Evaluation of QCNN-LSTM for Disability Forecasting in Multiple Sclerosis using Sequential Multisequence MRI

John D. Mayfield, M.D., M.Sc. [1,2], Issam El Naqa, Ph.D. [2.3]

1 University of South Florida Morsani College of Medicine

2 University of South Florida Department of Engineering – Medical Engineering

3 H. Lee Moffitt Cancer Center Department of Machine Learning

ABSTRACT

**Introduction**

Quantum Convolutional Neural Network (QCNN) - Long Short-Term Memory (LSTM) models were studied to provide sequential relationships for each timepoint in MRIs of patients with Multiple Sclerosis (MS). In this pilot study, we compare three QCNN-LSTM models for binary classification of MS disability benchmarked against classical neural network architectures. Our hypothesis is that quantum models will provide competitive performance.

**Methods**

Matrix Product State (MPS), Reverse Multistate Entanglement Renormalization Ansatz (MERA), and Tree-Tensor Network (TTN) circuits were paired with LSTM layer to process near-annual MRI data of patients diagnosed with MS. These were benchmarked against a Visual Geometry Group (VGG16)-LSTM and a Video Vision Transformer (ViViT). Predicted logits were measured against ground truth labels of each patient's Extended Disability Severity Score (EDSS) using binary cross-entropy loss. Training/Validation/Holdout Testing was partitioned using 5-fold cross validation with a total split of 60:20:20. Levene's test of variance was used to measure statistical difference and Student's t-test for paired model differences in mean.

**Results**

The MPS-LSTM, Reverse MERA-LSTM, and TTN-LSTM had holdout testing ROC-AUC of 0.70, 0.77, and 0.81, respectively (p-value 0.915). VGG16-LSTM and ViViT performed similarly with ROC-AUC of 0.73 and 0.77, respectively (p-value 0.631). Overall variance and mean were not statistically significant (p-value 0.713), however, time to train was significantly faster for the QCNN-LSTMs (39.4 seconds per fold vs. 224.3 and 217.5, respectively, p-value <0.001).

**Conclusion**

QCNN-LSTM models perform competitively compared to their classical counterparts with greater efficiency in train time. Clinically, these can add value in terms of efficiency to time-dependent deep learning prediction of disease progression based upon medical imaging.

# 1 Introduction

## 1.1 Time Dependent Machine Learning

Temporal relationships in data can provide critical information in experimental predictions of future outcomes. This is prevalent in areas of perceived uncertainty spanning from medical disease progression, stock market performance, and adverse weather conditions. This has traditionally been modeled using classical machine learning (ML) architectures ranging from Recurrent Neural Networks (RNNs) to the most recent innovation, transformers. In the medical imaging space, however, there have been few studies to predict disease progression based upon temporal data from imaging modalities such as Computed Tomography (CT) (Jianbo, 2022; Kim, 2021; Lucas, 2018), Magnetic Resonance Imaging (MRI) (Wang, 2024, Yan, 2022), and functional MRI (fMRI) (Yoon, 2020).

Given a time series $X(T) = \{x(1), x(2), ..., x(t)\}$, $t \in \mathbb{N}$ and the respective labels $y_i \in Y = \{1 \leq k \leq K\}$ for $K$ classes, the accurate prediction of $\tilde{y}_i \in \tilde{Y}$ where $\tilde{Y}$ is the set of labels on unseen testing time-series data, $\tilde{x}$, is the objective of some function $f$ such that $\tilde{y}_i = f(\tilde{x}, \theta^*)$ where $\theta^*$ is the parameter resulting from the minimization of a cost function, $C(\theta)$ such that $\theta^* = \text{argmin}_\theta C(\theta)$. Specific to medical imaging classification, initial image inputs can undergo dimensionality reduction such as Principal Component Analysis (PCA) or Uniform Manifold Approximation and Projection (UMAP), a convolutional neural network (CNN), or more recently, a vision transformer (ViT) by Dosovitskiy, et al. (https://arxiv.org/abs/2010.11929) to provide temporal encoding for each time step. These inputs are subsequently passed into an architecture that can provide temporal memory as seen in Recurrent Neural Networks (RNNs) such as a Gated Recurrent Unit (GRU) and Long Short-Term Memory (LSTM). In our study, we will also explore the tokenization of time units known as tubelet encoding within a novel Video Vision Transformer (ViViT) first introduced by Arnab, et al. (https://arxiv.org/abs/2103.15691). In prior studies, we have built upon an existing body of research to further explore the utility of time-dependent deep learning models in predicting progression of disease based upon medical imaging when presented in a sequence to both CNN-RNN and more novel ViT-LSTM and ViViT. Initial studies have shown promising results, however, more investigation using larger populations is warranted. Given the continued evolution of near-term quantum computing and potential improved efficiency, we sought to test the current hypothesis whether a Quantum Convolutional Neural Network (QCNN)-based architecture will perform competitive to their classical counterpart.

## 1.2 Quantum Advantage

The current climate for quantum computing still exists in the paradigm of Noisy Intermediate Scale Quantum (NISQ) where the hardware is costly, noisy, and expensive. However, there is an evolving body of literature describing how quantum computing methods are quickly demonstrating non-inferiority to classical methods, especially in the field of Quantum Machine Learning (QML) as first described by Wittek (2014). Specifically, the ability of QML architectures to manipulate the unique properties of quantum states to represent features and manipulate them in infinite-dimensional Hilbert space over the traditional Euclidean real-valued space. Theoretically, this can provide both real and imaginary values for each of the given input features, potentially doubling the number of trainable parameters with the same sample size requirements. In a branching architecture as seen in a feed forward network as described by Grant et al. (2018), each of the quantum states could be updated by weights for each layer, as well as subsequent pooling until a final convolutional output is calculated, thus mirroring a traditional CNN. As described by Pesah et al. (2021), this updating provides not only performance efficiency with $O(\log(n))$ runs instead of $O(n)$, but also provides a valuable connection to tensor networks, allowing for readily deployable Quantum-Classical (Q-C) architectures for machine learning, especially in time-series image classification.

1.3 Proposed Work

We propose to compare three different VQC-based architectures with an LSTM which are benchmarked against more traditional architectures including a Visual Geometry Group (VGG)16-LSTM and state of the art Video Vision Transformer (ViViT). These models will be tested on a two-site, single institutional dataset of sequential magnetic resonance imaging (MRI) with and without gadolinium contrast of the cervical spine in patients diagnosed with Multiple Sclerosis (MS). The models will predict binary classification of each patient's extended disability severity score (EDSS), a universal scoring system used by Neurologists and Neuroimmunologists to objectively measure and record disability. The EDSS, in combination with other symptoms and laboratory values, helps clinicians customize their existing arsenal of therapies. The original study population was identified by Neurologists with stratification of their disease in hopes to provide a prognostic tool to help not only their clinical decision-making, but also their patients' life planning as MS can be a severely debilitating disease in formative years. As there are current early works in prediction of breast cancer in mammography (Chan, 2020; Chang, 2020), and pathology on chest x-rays (Alshamrani, 2022; Bhattacharjya, 2023; Jia, 2022; Kumar, 2022; Sinwar, 2022), this is the first study of longitudinal disease prediction using quantum deep learning methods.

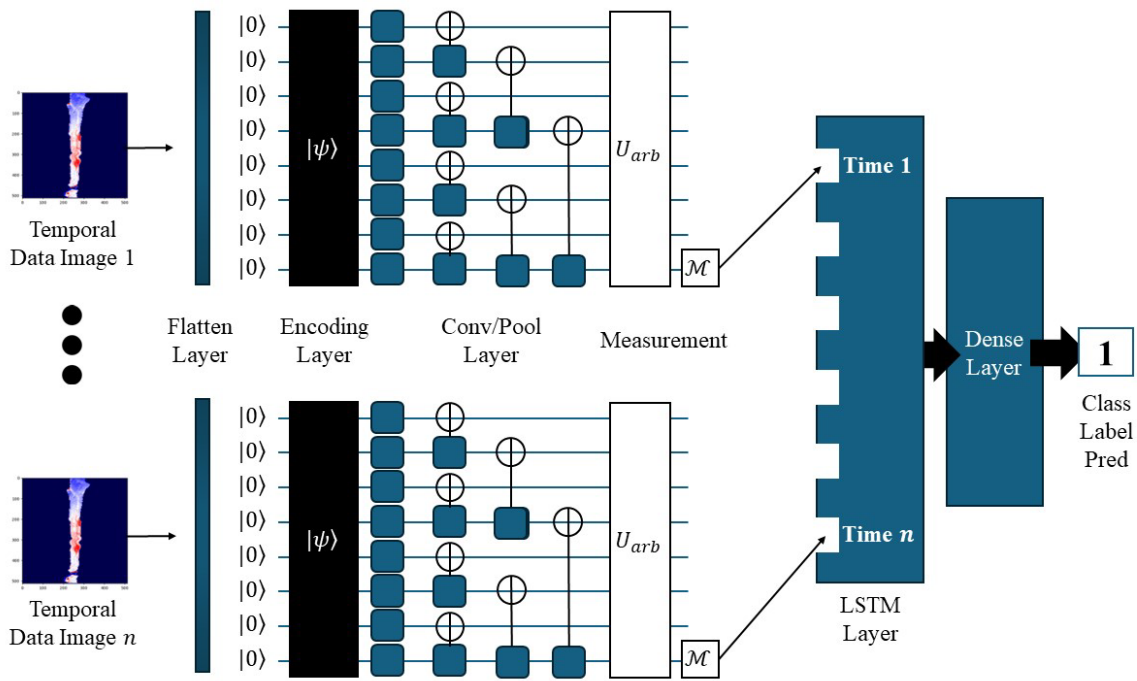

**Fig 1.** Generalized representation of the QCNN-LSTM architecture we created for our study. Each frame of a video input represents a single timepoint multisequence contrast-enhanced MRI which is flattened into size $2^{16}$ followed by Amplitude encoding. Quantum convolutional and pooling layers process the input through and arbitrary unitary "Dense" layer before the measured observable. This measurement is then passed to the respective temporal input in the LSTM layer. This is repeated for each frame and the final classical Dense layer provides the final binary classification.

## 2 Theory

### 2.1 Classification and Convolution

In a traditional CNNs, the initial step is a dot product of a subset of the image $x_i \in X$ and weight matrices $W^k$ where $k$ represents the hidden layer appropriated to that weight matrix, such that a linear combination of the inputs is $y_i^\ell = \{w_0^\ell x_{i,j} + w_1^\ell x_{i+1,j} + w_2^\ell x_{i,j+1} \cdots w_n^\ell x_{i+n,j+n}\}$ where $\ell \in \mathbb{N}$ represents the hidden layer. Similar to traditional CNNs, QCNNs ideally provide sequential convolutions on a feature space from a given input, but the representation of these features is projected onto an infinite-dimensional Hilbert space instead of a three-dimensional Euclidean space. Based upon the embedding strategy, each qubit represents some component of the input feature space. This signal is then passed through a variety of gates that rotate the state within the Bloch sphere based upon the axes x, y and z, each with parameters $\theta$. To mirror a CNN feed-forward network, we initiated the initial gate rotations with updated weights serving as the rotation values for subsequent gate rotations. As each of these gates are unitary such that $U_j(\theta_j) = e^{-i\frac{\theta_i}{2}P_i^\dagger}$ with $P_i^\dagger$ as a Pauli operator, the overall transformation of the initial input can be summarized as $U(x_i, \theta) = U_n(\theta_n) \cdot U_{n-1}(\theta_{n-1}) \cdots U_1(\theta_1) \cdot U_0(x_i)$. For application to a quantum circuit, the entire function is represented as $f(x_i, \theta) = \langle\psi_{i-1}|U_i^\dagger(\theta_i)\hat{B}_{i+1}U_i(\theta_i)|\psi_{i-1}\rangle = \langle\psi_{i-1}|\mathcal{M}_{\theta_i}(\hat{B}_{i+1})|\psi_{i-1}\rangle$, where $\hat{B}_{i+1}$ represents the measurement operator on the final qubit given a multi-gate system. The respective gradient would then be calculated as $\nabla_{\theta_i} f(x, \theta) = \langle\psi_{i-1}|\nabla_{\theta_i}\mathcal{M}_{\theta_i}(\hat{B}_{i+1})|\psi_{i-1}\rangle$.

#### 2.1.1 Background

Variational Quantum Classifiers (VQCs) are a form of parametrized quantum circuits that embed classical data into quantum states utilizing encoding methods such as Basis embedding, Angle embedding, Quantum, embedding and Amplitude embedding to name a few. This quantum representation of classical data is then passed through circuit which exists in a fixed initial state with gates (linear operators) that contain trainable parameters $\theta$ (usually rotation angles depending on the construct), and a final observable measurement that translates the quantum state(s) back into classical data representing a binary outcome probability. Specifically, these expectation values are calculated via a final nonlinear activation that computes the absolute square of the final two qubits and linearly combines them into the scalar probability. This expectation value can be estimated as such: $\langle\psi|O|\psi\rangle = \sum_{m,n=0}^{2^N-1}\langle\psi|m\rangle\langle m|O|n\rangle\langle n|\psi\rangle$ where $m, n$ refer to the computational basis states of two qubits, $|m\rangle$ and $|n\rangle$. Regarding the trainable parameters $\theta$, it is important to distinguish that these are not the same as weights in classical neural networks $w_{ij} \in W$, but rather parametrize these akin to $w_{ij}(\theta)$ (Schuld, 2019).

#### 2.1.2 Matrix Product State

We initially utilized the Matrix Product State (MPS) VQC originally described by Guifre Vidal 2003 (https://arxiv.org/abs/quant-ph/0301063). We implemented a similar architecture introduced by Huggins, et al. (https://arxiv.org/abs/1803.11537) with sequential product states of $n$ qubits in a cascade resulting in a final measurement of the observable for each frame (timepoint) of the patient's data. MPS is a concept established in an effort to improve the density-matrix renormalization group method (DMRG) which has historically been an effective method in static and dynamic simulation of quantum lattice systems (Schollwock, 2011). Generally speaking, MPS provides a "patch" based approach to complex quantum data such as a quantum many-body state that allows processing of smaller components with the resultant product passed to the next state:

$$|\Psi\rangle = \sum_{\{s\}} \text{Tr}\left[A_1^{(s_1)} A_2^{(s_2)} \cdots A_N^{(s_N)}\right] |s_1 s_2 \cdots s_N\rangle$$

where $|\Psi\rangle$ is the MPS of complex square matrices $A_N^{(s_N)}$ for indices $s_i$ representing the states within the computational basis. A full exploration of the many manifestations of MPS use in QML is beyond the scope of the paper; however, this is an evolving area of research in increasing quantum computational efficiency. The graphic of our specific architecture used with 16 qubits is shown later in Figure S1 where paired sequential U3-IsingXX-IsingYY-IsingZZ-U3 gates. A curtailed example is shown below using 4-qubits in Figure 2a.

2.1.3 Reverse MERA

Multiple-scale Entanglement Renormalization Ansatz (MERA) is a variational circuit that provides feature extraction of data using sequential unitary and isometry layers which introduce other qubits resulting in a tensor network representation of the many-body wavefunctions. For implementation as a QCNN, the MERA is reversed where entangled qubits are reduced step-wise with passed representations until a single qubit remains, thus producing a measurable outcome. As such, a given input state following encoding $|\psi\rangle$ can theoretically produce the corresponding label. A curtailed example is shown below using 4-qubits in Figure 2b.

2.1.4 Tree Tensor Network (TTN)

First described by Grant et al. (2018), the Tree Tensor Network (TTN) mimics classical feed-forward neural network architectures with stepwise halving of filtered data by pooling functions. Specific to the quantum circuit, the pooling consists of discarding qubit outputs from each unitary between paired nearest neighbor qubits which continues until a single qubit is left on which a measurable outcome is determined. This single-qubit expectation can be described by the following equation (Grant, 2018):

$$\mathcal{M}_\theta(\psi^a) = \langle \psi^a | \hat{U}_C^\dagger(\{U_i(\theta_i)\}) \hat{\mathcal{M}} \hat{U}_C(\{U_i(\theta_i)\}) | \psi^a \rangle$$

where $\hat{U}_C(\{U_i\})$ represents a quantum circuit composed of unitary $U_i(\theta_i)$ for parameters $\theta$ and $\mathcal{M}$ represents the measurement operator acting on the final qubit of the circuit. Similar to CNNs, TTNs can be optimized using stochastic gradient descent. A curtailed example using 4-qubits is shown in Figure 2c.

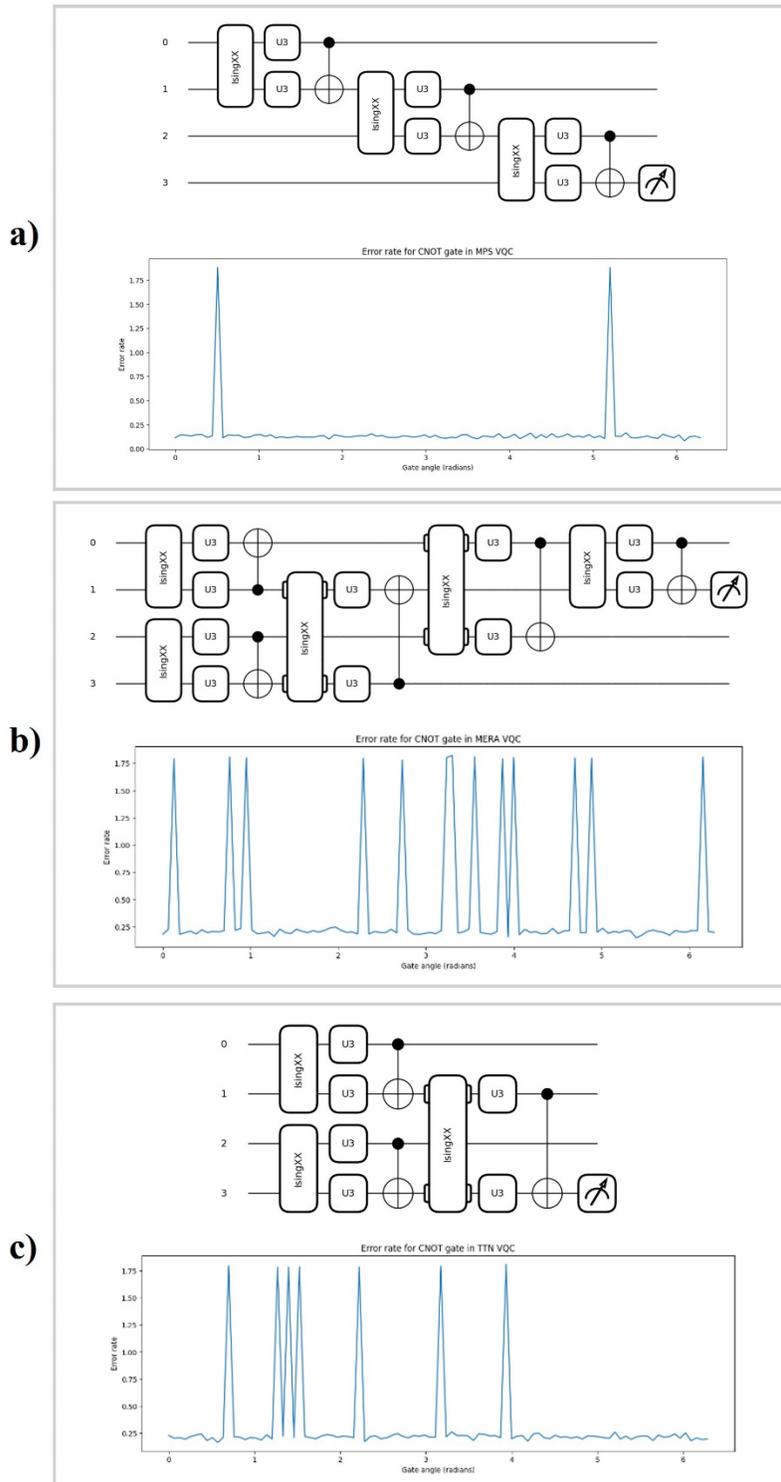

**Figure 2.** Toy examples of the **(a)** Matrix Product State, **(b)** Multi-state Entanglement Renormalization Ansatz, and **(c)** Tree Tensor Network-QCNN architecture with pooling via CNOT gates on every other qubit. Output from the measurement impacts the state of the remaining qubit which undergoes continued

convolution with subsequent qubits until a final measured observable is passed to the classical model. Each model shows the respective error rates for the CNOT gates in each circuit, calculated as the difference between zero shots and 1000 shots. Circuit diagrams were created in PennyLane.

2.2 Encoding

We selected Amplitude Encoding as our embedding strategy to optimize the maximal image size (256x256) and minimal qubits (16). This is in contrast to prior discussion by Grant et al., we found that Qubit encoding, while ideal, required prior dimensionality reduction of the image features to 16 components prior to encoding which may have detrimental information loss of high-resolution images as seen in MRI. PCA uses linear techniques which may lose high-frequency information, and UMAP uses nonlinear techniques which may degrade small details in the image (Velliangiri and Alagumuthukrishnan, 2019). Amplitude Encoding or Embedding is a method of encoding classical data of dimension $N = 2^n$, specific to our study, a grayscale image of size $NxN$ with normalized grayscale values $n_i \in [0,1]$. This data is subsequently then transformed into quantum state amplitudes $|\phi(x)\rangle$. Specifically, a nonlinear activation function is applied to an input vector $\mathbf{v} \in \mathbb{R}^N$ with each vector $v_i \to \varphi(v_i)$ which is first normalized, providing a distinction from feed-forward neural network activation, with each element in the vector serving as the amplitude for each respective quantum state. As a toy example, consider a two-dimensional vector with coordinate points $\{(a,b),(c,d)\}$:

$$v_i = [a,b,c,d]$$

$$v_i' = \frac{1}{\sqrt{a^2+b^2+c^2+d^2}}[a,b,c,d]$$

$$|v_i'\rangle = \frac{1}{\sqrt{a^2+b^2+c^2+d^2}}(a|00\rangle + b|01\rangle + c|10\rangle + a|11\rangle)$$

where $|v_i'\rangle$ is the resultant amplitude vector with the values $a,b,c,d \in \mathbb{R}$ as the respective amplitudes for each of the four qubits in this example. More generally, this can be represented as

$$|\phi(x)\rangle = \frac{1}{\|x\|}\sum_{i=1}^{N} x_i|i\rangle$$

where $i \in \mathbb{N}$ representing the respective qubit and $|i\rangle$ the *ith* computational basis state.

**3 Time-Dependent Quantum Deep Learning Methods**

3.1 Ansatz

Each of the base architectures detailed above were modified to include sequential U3 gates, a single qubit unitary. The U3 provides three distinct qubit rotations: polar $\theta$, azimuthal $\phi$, and quantum phase $\delta$. The last entity is a rotation about the z-axis following initial y-rotation and is also designated by the symbol $\lambda$. Of note, quantum phase may also be referred to as the phase factor.

$$U_3(\theta,\phi,\delta) = \begin{bmatrix} \cos\left(\frac{\theta}{2}\right) & -e^{i\delta}\sin\left(\frac{\theta}{2}\right) \\ e^{i\phi}\sin\left(\frac{\theta}{2}\right) & e^{i(\phi+\delta)}\cos\left(\frac{\theta}{2}\right) \end{bmatrix}$$

The U3 can also be conceptualized as single-qubit R(Rot) and R(Phase Shift) such that $U_3(\theta, \phi, \delta) = R_\phi(\phi + \delta)R(\delta, \theta, -\delta)$. Following the U3 gate, sequential IsingXX, IsingYY, and IsingZZ gates were implemented on each qubit. These gates provide two-qubit tensor products of two Pauli-X $\begin{pmatrix} 0 & 1 \\ 1 & 0 \end{pmatrix}$, Pauli-Y $\begin{pmatrix} 0 & -i \\ i & 0 \end{pmatrix}$, or Pauli-Z $\begin{pmatrix} 1 & 0 \\ 0 & -1 \end{pmatrix}$ matrices.

$$IsingXX(\phi) = e^{-i\left(\frac{\phi}{2}\right)(X \otimes X)} = \begin{bmatrix} \cos\left(\frac{\phi}{2}\right) & 0 & 0 & -i\sin\left(\frac{\phi}{2}\right) \\ 0 & \cos\left(\frac{\phi}{2}\right) & -i\sin\left(\frac{\phi}{2}\right) & 0 \\ 0 & -i\sin\left(\frac{\phi}{2}\right) & \cos\left(\frac{\phi}{2}\right) & 0 \\ -i\sin\left(\frac{\phi}{2}\right) & 0 & 0 & \cos\left(\frac{\phi}{2}\right) \end{bmatrix}$$

$$IsingYY(\phi) = e^{-i\left(\frac{\phi}{2}\right)(Y \otimes Y)} = \begin{bmatrix} \cos\left(\frac{\phi}{2}\right) & 0 & 0 & -i\sin\left(\frac{\phi}{2}\right) \\ 0 & \cos\left(\frac{\phi}{2}\right) & -i\sin\left(\frac{\phi}{2}\right) & 0 \\ 0 & -i\sin\left(\frac{\phi}{2}\right) & \cos\left(\frac{\phi}{2}\right) & 0 \\ -i\sin\left(\frac{\phi}{2}\right) & 0 & 0 & \cos\left(\frac{\phi}{2}\right) \end{bmatrix}$$

$$IsingZZ(\phi) = e^{-i\left(\frac{\phi}{2}\right)(Z \otimes Z)} = \begin{bmatrix} e^{-i\left(\frac{\phi}{2}\right)} & 0 & 0 & 0 \\ 0 & e^{i\left(\frac{\phi}{2}\right)} & 0 & 0 \\ 0 & 0 & e^{i\left(\frac{\phi}{2}\right)} & 0 \\ 0 & 0 & 0 & e^{-i\left(\frac{\phi}{2}\right)} \end{bmatrix}$$

In the Ising-XX and Ising-YY matrices, these are 4 x 4 matrices representing combined space state of the two qubits where $\cos\left(\frac{\phi}{2}\right)$ represents that the state in this position remains unchanged, and $-i\sin\left(\frac{\phi}{2}\right)$ represents an entangling operation with the negative sign indicating the direction of rotation within a Bloch sphere. In the Ising-ZZ matrix, the sign preceding the exponent indicates the direction of phase shift. Additionally, the remaining zeros outside of the diagonal represent that there is no mixing of states, simply phase shifts.

3.1.1 Convolutional Layer

Each qubit is initiated with randomized weights for the U3 gate angle rotations, $\theta, \phi$ and $\delta$, respectively. Each subsequent IsingXX, YY, and ZZ gates for paired qubits $n$ and $n + 1$ with weights $w_{ij}$. Subsequent U3 gates receive the updated rotation values much as weights do within a CNN convolutional layer.

3.1.2 Pooling Layer

After each sequence of U3, IsingXX, IsingYY, IsingZZ, and subsequent U3 gate, a pooling operation reduces the number of qubits by 2 for each layer via Control-NOT (CNOT) gates, as described in Section

2.1.4. where the first qubit of the pair is a control and the second qubit, a target. The resultant measurement can take the values of -1 or 1 which then impact the rotation of the next qubit. Concurrently, this halves the number of qubits until a single qubit remains which undergoes a Dense layer activation via an Arbitrary Unitary gate.

3.1.3 Dense Layer

A final gate we used prior to measurement of the observable was an Arbitrary Unitary gate, $V = \begin{pmatrix} v_{00} & v_{01} \\ v_{10} & v_{11} \end{pmatrix}$ that serves as a Dense layer that would be seen in a CNN where two qubits are parametrized by $4^n - 1$ independent real parameters. The important component is that this gate is a unitary operator thus preserving the probability of all measurement outcomes by preserving the state norm. The quantum convolutional layer is now complete, and each frame input is fed sequentially to the LSTM layer in the classical half of the architecture.

3.2 Q-C with LSTM Layer

Time-dependent deep learning mandates the use of a memory strategy to establish relationships between timepoints and subsequent prediction of future events. Based upon a sizeable body of research, LSTMs provide a more robust memory passage layer to layer than GRUs when it comes to greater length of time periods, as well as retention of a greater number of parameters and operations. Additionally, LSTMs are more sensitive to hyperparameter tuning thus allowing a greater level of fine tuning within the models (Greff, 2015: https://arxiv.org/abs/1503.04069). In prior studies by our team, we established the superiority of CNN-LSTM over CNN-GRU specifically in MRI disease prediction from sequential patient data. Each of the observable measurements from the respective frames (timepoints) for each patient are fed sequentially into a traditional LSTM with 1024 layers and a dropout of 0.5.

3.3 Cost Function

A potential advantage in quantum circuits is the utility of parameter-shift rules by Mitarai (2018) further expanded upon by Schuld (2019) which utilize partial differential equations (PDEs) to estimate the exact value of a gradient theoretically. Generally, these PDEs provide derivatives on shifted trigonometric equations, e.g. $\cos(\theta + \frac{\pi}{4})$ which allow computation of the original entity and its derivative without changing the architecture. This is further detailed in Section 2.1. Specific to our study, we selected binary crossentropy loss to measure variance between the calculated expected value and the true label in order to parallel the classical architectures by which the QCNN-LSTM models were benchmarked. From this paradigm, probabilities of computational basis states can be optimized with a single-qubit measured observable. The formula can be represented by the following relationship

$$C(\theta) = \sum_{i-1}^{N} y_i \log(\Pr[\psi_i(\theta) = 1]) + (i - y)\log(\Pr[\psi_i = 0])$$

where a label, $y_i \in [0,1]$ and $\Pr[\psi_i(\theta) = y_i]$ is the probability of measuring the computational base state, given a binary classification in a QCNN architecture.

**4 Experiments**

4.1 Data

Data collection was compliant with Health Insurance Portability and Accountability Act (HIPAA) as well as the World Medical Association (WMA) Declaration of Helsinki. The study was approved by the institutional review board (IRB) with waiver of the requirement of informed consent due to the retrospective nature of the study. A total of 2232 patients diagnosed with MS between the years of 2006 and 2023 were identified by the USF Health Neuroimmunology department in collaboration with Tampa General Hospital (TGH). To establish time-dependency of disease, inclusion criteria required at least two contrast-enhanced multisequence MRIs. This yielded a total of 703 patients. Average patient age was 46.6 (IQR 37-56) with average EDSS of $3.5\pm2.5$. Per the Neuroimmunologists caring for these patients, binary classification of EDSS was subdivided into those with scores of 0 to 3 (no need for assistive devices with minor comorbidities) and those 4 and above (require assistive devices to ambulate with varying moderate to severe comorbidities). Data was partitioned into an 60:20:20 split for training, validation, and holdout testing.

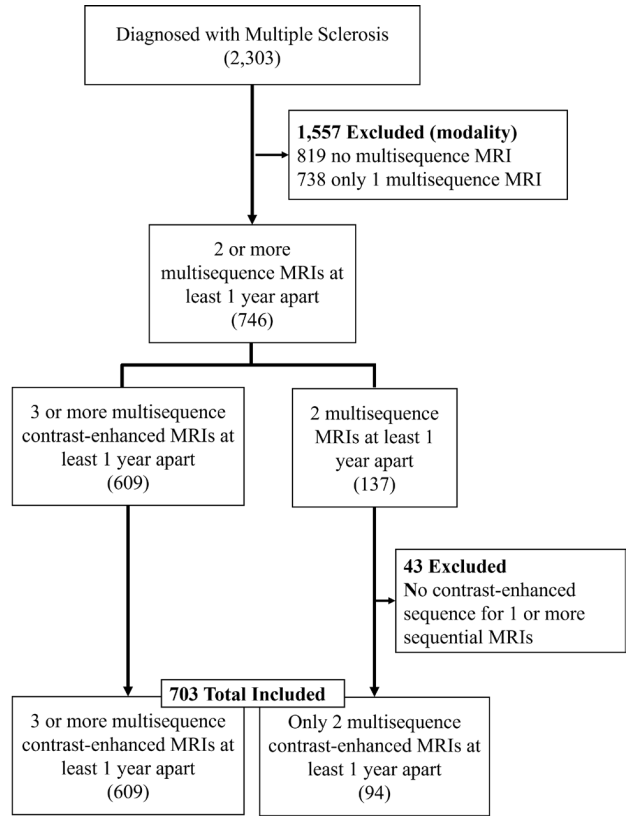

### 4.1.1 Data Partitioning

The data cohort was randomly split into 80:20 training and hold-out testing prior to stratified k-fold cross-validation (k = 5). Hold-out testing was randomly initialized three (3) times with an average performance reported.

### 4.2 Impact of Qubit Size

For our experiment, we took inspiration from Grant, et al. in the utilization of TTN and MERA with the additional investigation of MPS as QCNN backbones for the Q-C models. Initial testing of the impact of qubit size (and image resolution) demonstrated a increased performance at 16-qubits over 14 and 12 (Figure 3). We attempted 18 qubits with image size 512x512, however, this required memory in excess of the A100 GPU capacity.

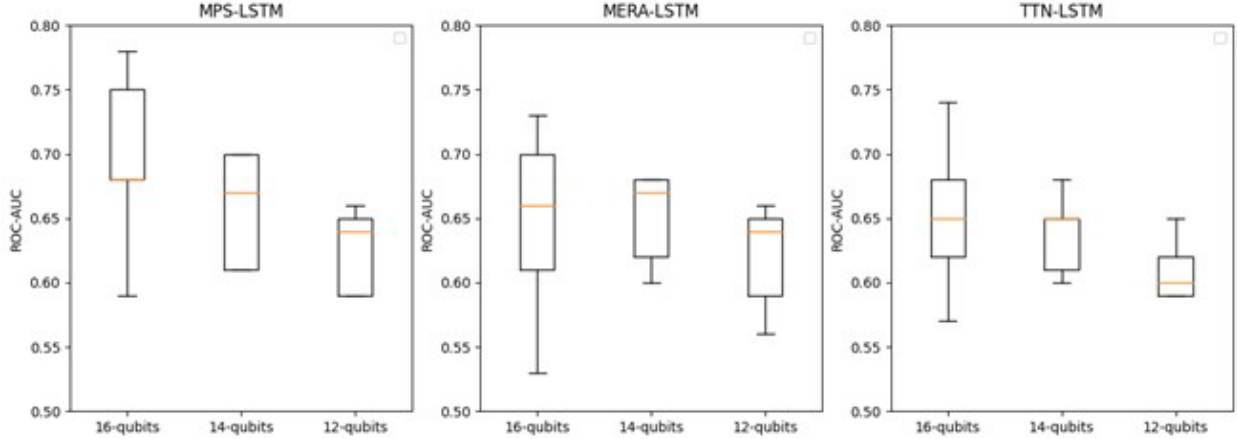

**Figure 3**. Comparison of the difference in general training performance using a decreasing number of qubits.

4.3 QCNN-LSTM Results

To measure the variance of the different QCNN-LSTM models in comparison to the VGG16-LSTM and ViViT, we used a Levene's Test for Equality of Variances between models resulting in a p-value of 0.713 amongst all models with a VQC backbone, as well as 0.713 for all models including classical architectures. Additionally, a paired Student's T-test was used to evaluate differences in mean (Table 1).

**Table 1.** .VQC-LSTM Performance across all classes. Validation receiver operator curve – area under the curve (ROC-AUC) was averaged over 5-fold stratified cross-validation for each model. Holdout testing accuracy (Acc), ROC-AUC are reported separately to demonstrate congruency in performance across the classes. The reported precision, recall, and harmonic mean (F1-score) are weighted averages across the two class labels.

| VQC Ansatz | Validation AUC (95% CI) | Holdout Test Acc | Holdout Test AUC | Precision (Avg) | Recall (Avg) | F1-score (Avg) |
|---|---|---|---|---|---|---|
| *MPS-LSTM* | 0.66 (0.52-0.78) | 0.70 | 0.70 | 0.70 | 0.70 | 0.71 |
| *MERA-LSTM* | 0.70 (0.50-0.80) | 0.76 | 0.77 | 0.78 | 0.76 | 0.76 |
| *TTN-LSTM* | 0.71 (0.55-0.84) | 0.76 | 0.81 | 0.75 | 0.75 | 0.75 |
| | | | *p-value 0.915* | | | |

**Table 2.** Individual Class Metrics. Scores below are split into mild / severe disability for the three different VQC-LSTMs, respectively.

| VQC Ansatz | Precision | Recall | F1-score |
|---|---|---|---|
| *MPS-LSTM* | 0.70 / 0.70 | 0.85 / 0.49 | 0.77 / 0.58 |
| *MERA-LSTM* | 0.86 / 0.67 | 0.70 / 0.84 | 0.77 / 0.74 |
| *TTN-LSTM* | 0.79 / 0.70 | 0.79 / 0.70 | 0.79 / 0.70 |
| | *p-value 0.223* | | |

**Table 3.** Confusion matrices on the holdout test set across the three QCNN-LSTM models.

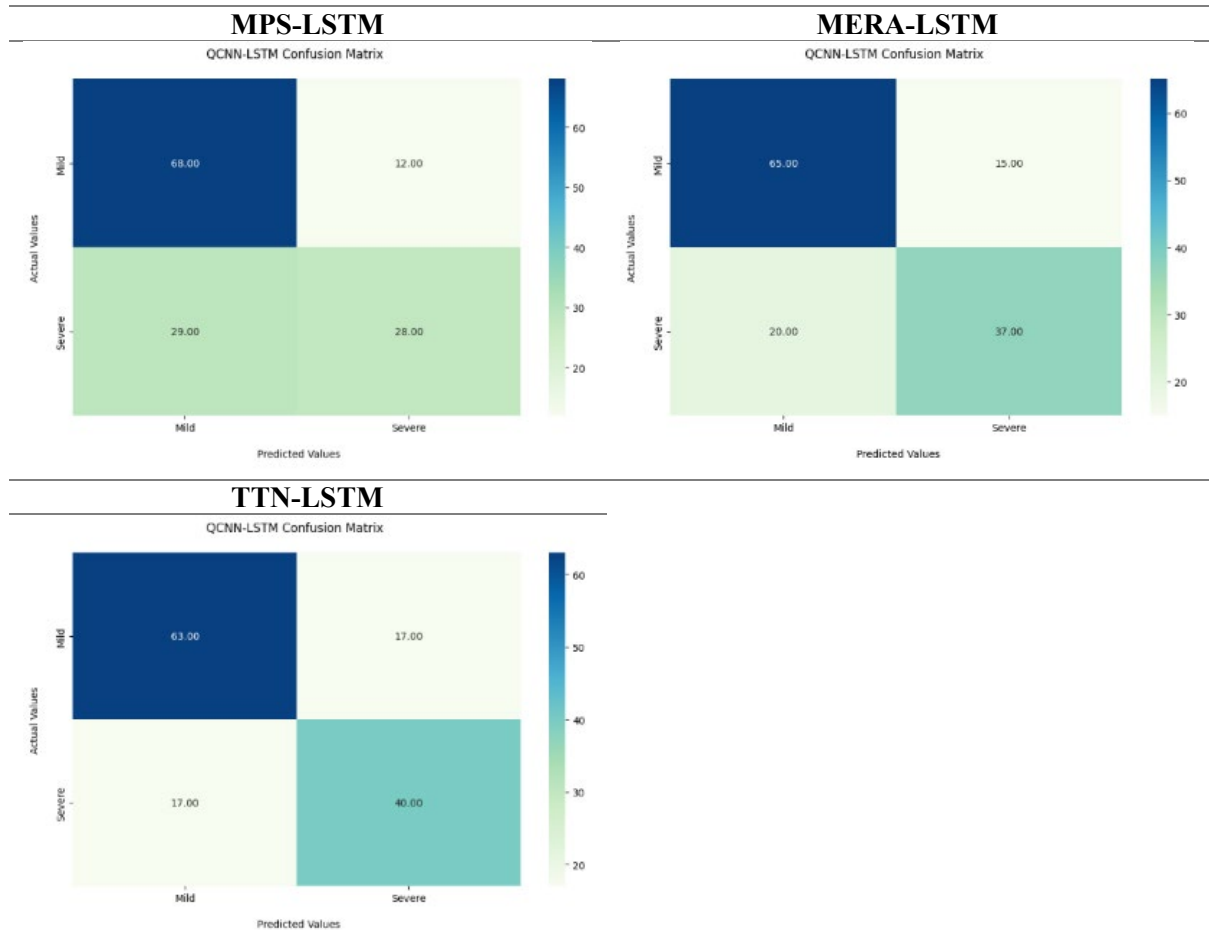

4.4 Noise and Error Modeling

We introduced simulated noise and error with the following tools:

1. Pennylane Depolarizing Channel (https://docs.pennylane.ai/en/stable/code/api/pennylane.DepolarizingChannel.html?highlight=depolarizing%20channel)
2. Amplitude Damping (https://docs.pennylane.ai/en/stable/code/api/pennylane.AmplitudeDamping.html?highlight=amplitude)
3. Phase Damping (https://docs.pennylane.ai/en/stable/code/api/pennylane.PhaseDamping.html?highlight=qml%20phasedamping#pennylane.PhaseDamping)
4. BitFlip (https://docs.pennylane.ai/en/stable/code/api/pennylane.BitFlip.html)

for each qubit with incremental increase from 0.0 through 1.0 in increments of 0.25. This allowed us to visualize effects of known full depolarizing channels (noise = 0.75) and uniform Pauli-error channels (noise = 1.0) We monitored stratified training folds for ROC-AUC loss and subsequent holdout testing ROC-AUC and F1-score degradation. Across all models and noise levels, there was a more pronounced visualized decrease in F1-score (p-value 0.025) despite a relatively stable ROC-AUC in the MERA- and TTN-based model (p-value 0.443). Of note, MPS remained relatively unaffected for both metrics.

**Table 4.** Comparative Effect of Varying Noise on ROC-AUC and F1-score on the holdout testing set.

|  | Noise Level | | | | |
|---|---|---|---|---|---|
|  | **0.00** No noise | **0.25** | **0.50** | **0.75** Full depolarizing channel | **1.00** Uniform Pauli error channel |
| **MPS-LSTM** | | | | | |
| *Test ROC-AUC* | 0.73 | 0.72 | 0.73 | 0.73 | 0.75 |
| *F1-Score* | 0.69 | 0.72 | 0.73 | 0.73 | 0.74 |
| **MERA-LSTM** | | | | | |
| *Test ROC-AUC* | 0.77 | 0.77 | 0.77 | 0.77 | 0.73 |
| *F1-Score* | 0.76 | 0.74 | 0.74 | 0.70 | 0.70 |
| **TTN-LSTM** | | | | | |
| *Test ROC-AUC* | 0.81 | 0.81 | 0.80 | 0.80 | 0.80 |
| *F1-Score* | 0.75 | 0.75 | 0.75 | 0.70 | 0.70 |

*ROC-AUC p-value 0.443*
***F1-Score p-value 0.025***

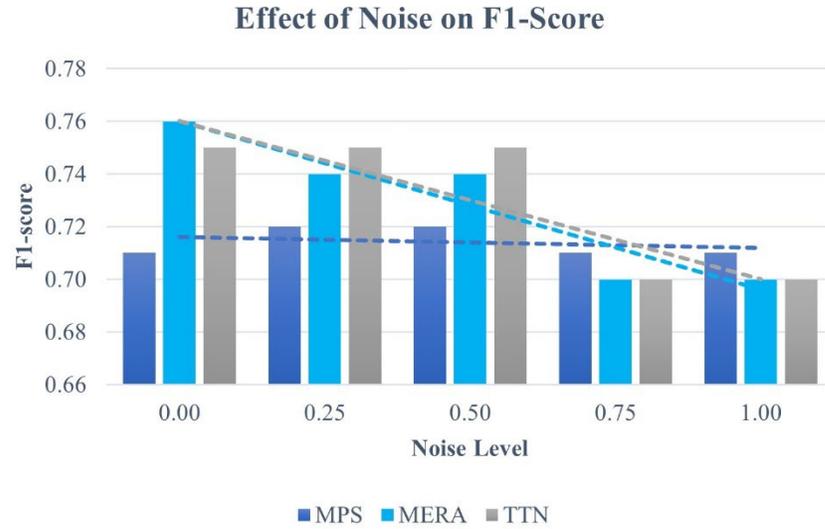

(a)

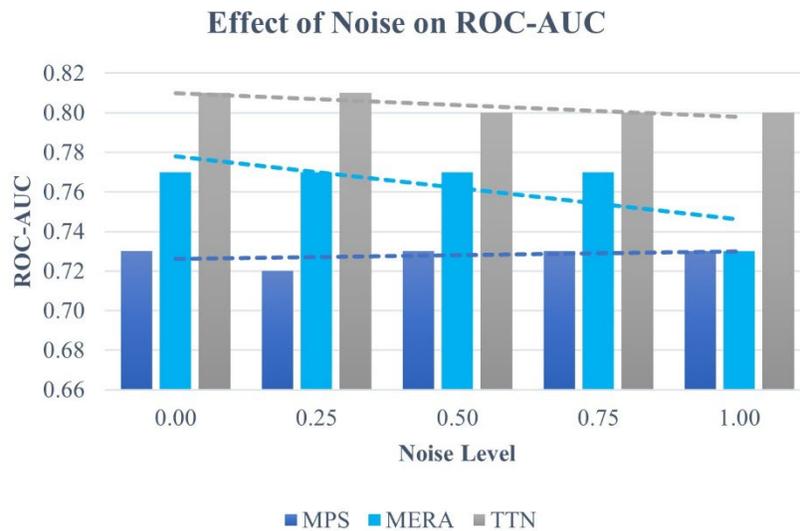

(b)

**Figure 4.** Graphical Representation of Effect of Varying Noise on (a) ROC-AUC and (b) F1-score in the holdout testing set.

4.5 Benchmarking Results

Prior studies investigating multiple sclerosis disability progression prediction was based upon exact EDSS and trinary classification tasks. This is the first study describing binary classification. Similar to these aforementioned studies, the ViViT demonstrates subtle improved performance over the CNN-LSTM architecture but not by a statistically significant amount. Overall, the performance of these classical models was not statistically significant from the QCNN-LSTM architectures (p-value 0.713).

**Table 5.** Classical Time-Dependent Model Performance across all classes. Validation receiver operator curve – area under the curve (ROC-AUC) was averaged over 5-fold stratified cross-validation for each model. Holdout testing accuracy (Acc), ROC-AUC are reported separately to demonstrate congruency in performance across the classes. The reported precision, recall, and harmonic mean (F1-score) are weighted averages across the two class labels. Binary classification confusion matrices are shown in (TABLE).

| Model | Avg. Validation AUC (95% CI) | Holdout Test Acc | Holdout Test AUC | Precision (WAvg) | Recall (WAvg) | F1-score (WAvg) |
|---|---|---|---|---|---|---|
| *VGG-LSTM* | 0.66 (0.56-0.73) | 0.70 | 0.73 | 0.70 | 0.69 | 0.69 |
| *ViViT* | 0.67 (0.59-0.75) | 0.74 | 0.77 | 0.74 | 0.74 | 0.74 |
| | *p-value 0.631* | | | | | |

**Table 6.** Individual Class Metrics. Scores below are split into mild / severe disability for the three different VQC-LSTMs, respectively.

| Model | Precision | Recall | F1-score |
|---|---|---|---|
| *VGG-LSTM* | 0.76 / 0.61 | 0.68 / 0.70 | 0.72 / 0.65 |
| *ViViT* | 0.75 / 0.74 | 0.85 / 0.60 | 0.80 / 0.66 |
| | *p-value 0.327* | | |

**Table 7.** Confusion matrices on the holdout test set across the three QCNN-LSTM models.

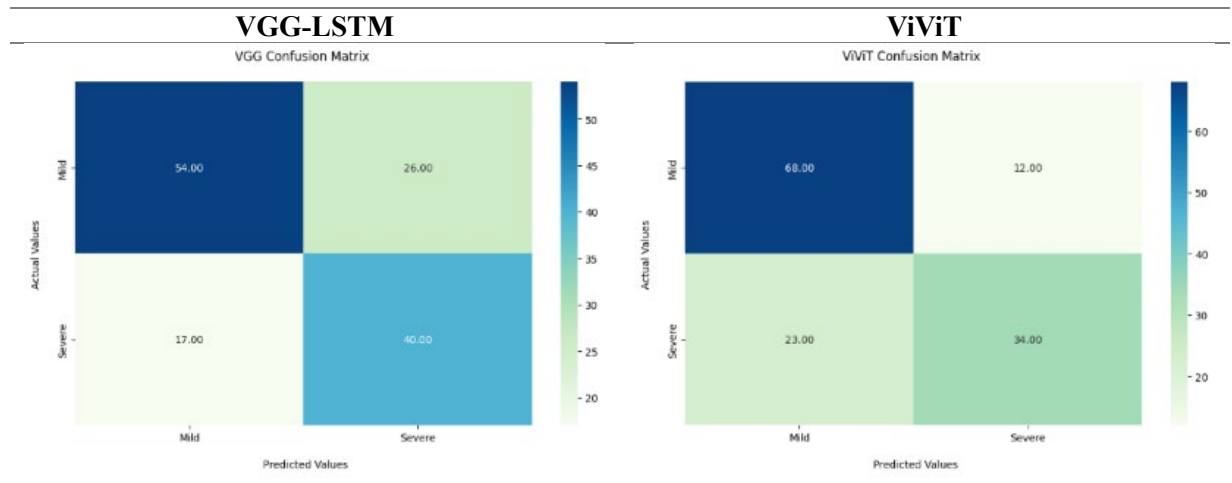

**Table 8.** Paired p-values between Models based upon Student's T-test. All reported p-values demonstrate the failure to reject the null-hypothesis such that there is no statistically significant difference between the means of the ROC-AUC between models.

|  | MPS-LSTM | MERA-LSTM | TTN-LSTM | VGG-LSTM | ViViT |
|---|---|---|---|---|---|
| MPS-LSTM | - | | | | |
| MERA-LSTM | 0.331 | - | | | |
| TTN-LSTM | 0.342 | 0.898 | - | | |
| VGG-LSTM | 0.212 | 0.962 | 0.823 | - | |
| ViViT | 0.539 | 0.529 | 0.568 | 0.321 | - |

4.6 Comparative Model Performance regarding Time to Train.

Measurements are by fold of training/validation with an early stopping parameter implemented based upon validation loss increase greater than 10 occurrences beyond 30 epochs on a small batch size of 4. This callback was uniform for all model training and validation as convergence was similar based on different learning rates per model.

**Table 9.** Relative Model Runtimes.

| | No. of Trainable Parameters $\theta$ (M=million) | Fold 1 Time (secs) | Fold 2 Time (secs) | Fold 3 Time (secs) | Fold 4 Time (secs) | Fold 5 Time (secs) | Mean Time (secs) |
|---|---|---|---|---|---|---|---|
| *QCNN-LSTM (Mean)* | 272M | 39.8 | 43.0 | 37.0 | 39.2 | 38.2 | **39.4** |
| *VGG-LSTM* | 23M | 218.9 | 272.8 | 223.4 | 204.7 | 201.5 | 224.3 |
| *ViViT* | 220M | 253.8 | 211.2 | 207.3 | 208.9 | 206.1 | 217.5 |
| | | | | | | | *p-value < 0.001* |

# 5. Discussion

## 5.1 Comparison to Classical Architectures

In this exploratory study, the QCNN-LSTM architectures demonstrated non-inferior performance to the classical time-dependent architectures, thus countering the existing zeitgeist that quantum models in their current NISQ designation are inferior to their classical counterparts. An interesting result from the study was the statistically significant run-time efficiency, averaging 39 seconds per training fold for the QCNN-LSTM models (p-value <0.001). Although conflicting to a general consensus, this study adds to a growing body of literature demonstrating the steadily improving quantum computing ecosystem (Yan, 2024). Specific to clinical application, quantum machine learning architectures may provide more efficient means of clinical utility in concert with classical models given the increasing access to cloud-based quantum devices with a growing number of qubits. As artificial intelligence becomes more pervasive throughout the medical sector, the continued evolution of time-dependent disease prediction and prognosis is paramount to optimized and personalized patient care models.

## 5.2 Bias and Error

There are certain limitations of this study that are important to discuss including the use of a simulator with estimated noise. The simulation of damping of qubit rotations or phase, or complete bit flips encountered in real NISQ devices may not provide the same consistent and competitive results as a consequence of lost convolutional information that would potentially establish vital relationships and identification of key features important to logit prediction. Additionally, we were unable to explore the effect of increasing qubit size beyond 16 given the memory constraints of the large number of parameter calculations and the higher resolution medical images. These future studies will also implement quantum error correction (QEC) methods to provide consistent, reliable models. Further investigation on real quantum hardware is a goal with investigation across varying qubit devices to measure performance variance between superconducting, ion trapping, diamond lattice, and photonic qubits. Of note, the MPS-LSTM demonstrated resilience to noise introduction in contrast to the MERA- and TTN-LSTMs and will certainly be a model of interest in future testing on real quantum hardware to see if its resilience is practical.

Beyond quantum hardware limitations, there is inherent bias within the dataset which may impact generalization to other populations. The data, while robust in size with near-equal class distribution, consisted of patients from a two-site single institution. While the demographic variation is present, the use of studying multiple sclerosis precludes to a class imbalance in race, and many times, gender assigned at birth as there is a greater prevalence of Caucasian women affected by the disease.

## 5.3 Future Works

As discussed above, there are many opportunities to expand this baseline study. We are currently working with two other academic institutions to implement transfer learning and external validation to test generalizability of the classical models. Based upon the findings from this pilot study, we are attempting to identify a similar scientific community to test external validation and evaluate them on various quantum computing architectures. From a clinical perspective, the next step is to implement multi-institutional observational trials without impact of the patients' care plans using a double-blind method. If the methodology is able to accurately predict a patient's future disability progression, we would provide this tool to clinicians to help them customize therapies to higher risk patients, as well as to patients to assist them in life planning, given the impact of MS during formative years.

## 6. Conclusions

Quantum computing can add value in terms of efficiency to time-dependent deep learning prediction of disease progression based upon medical imaging without significant classification performance difference from classical methods. There are many remaining obstacles to full practical implementation including testing on external validation sets with the use of real quantum hardware. This pilot study will serve as a foundation for us to develop further investigation of clinical application, specifically in the realm of MS disability prediction. It is our hope that this and other similar works will inspire others to push the envelope with time-dependent quantum ML solutions in order to provide more efficient, and one day more robust deep learning solutions to a variety of problems.

**Relevant Support**

This work was partly supported by National Institute of Health (NIH) grant R01-CA233487 and its supplement (CA233487-05S1).

**Conflict of Interest**

Issam El Naqa is on the scientific advisory of Endectra, LLC and co-founder of iRAI LLC, and act as deputy editor for the journal of Medical Physics and receives funding from NIH and DoD.

**Supplemental Material**

**Figure S1**. Global overview of the MPS circuit with 16 qubits to demonstrate cascading convolution of each pairwise qubit information to the next until a single qubit remains. While individual gates are not completely visualized, the purpose is to show the flow of qubit convolution and pooling.

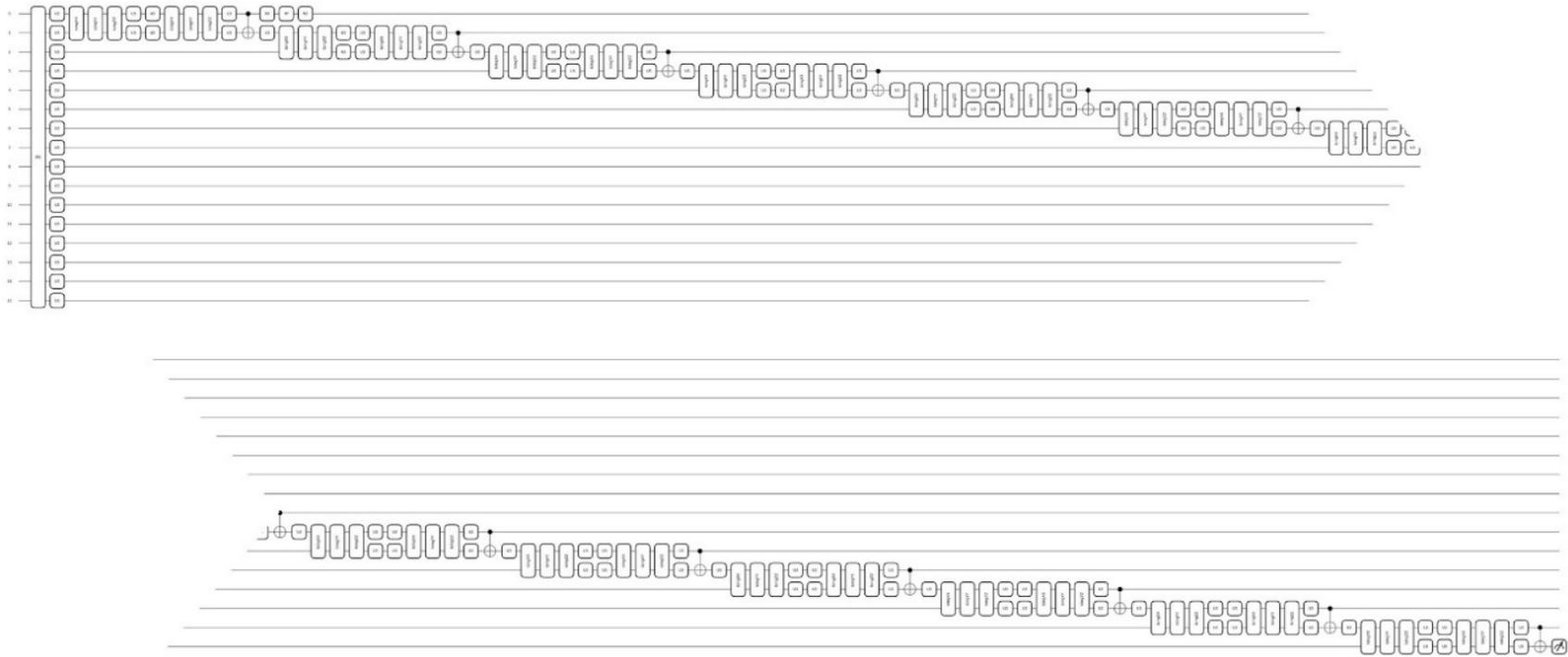

**Figure S2.** Global overview of the Reverse MERA VQC architecture with middle portion curtailed. This architecture is unique in its pattern of interleaving pairwise qubit entanglement with resultant single qubit output.

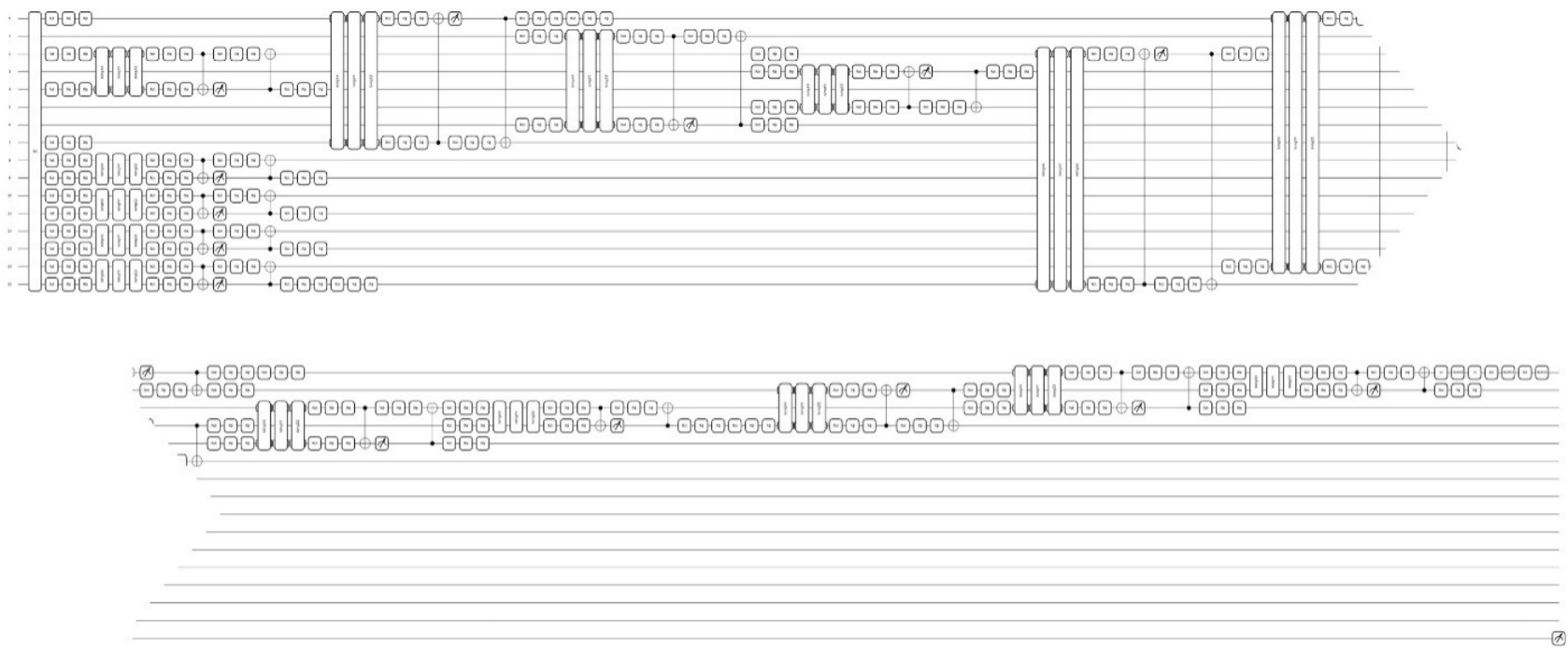

**Figure S3.** Global overview of the TTN VQC with 16 qubits demonstrating a pairwise qubit entanglement with pooling that mirrors a traditional convolutional neural network pattern. An arbitrary unitary operation is used prior to the measured observable to act as a Dense layer.

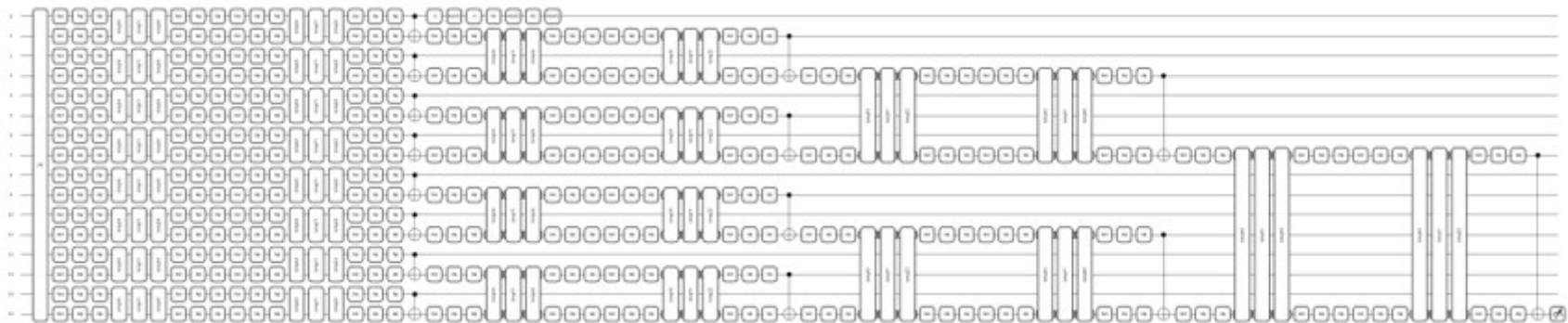